\documentclass[letterpaper, 10 pt, conference]{URL-ieeeconf}
\IEEEoverridecommandlockouts                \usepackage{makecell}

\usepackage{graphics}           
\usepackage{times}              
\usepackage{amsmath}            
\usepackage{amssymb}            
\usepackage{graphicx}
\usepackage{algorithm}
\usepackage[noend]{algpseudocode}
\usepackage{booktabs}
\usepackage{color}
\usepackage{multirow}
\usepackage{subcaption}
\usepackage{rotating}
\usepackage{cite}
\definecolor{instructioncolor}{rgb}{.5,.5,.5}

\def\eqref#1{(\ref{#1})}

\def\vsfig{\vspace{-0.3cm}}
\def\vstab{\vspace{-0.3cm}}

\newcommand{\rom}[1]{\uppercase\expandafter{\romannumeral #1\relax}}

\makeatletter
\usepackage{xspace}
\DeclareRobustCommand\onedot{\futurelet\@let@token\@onedot}
\def\@onedot{\ifx\@let@token.\else.\null\fi\xspace}

\makeatother

\usepackage{array}
\newcolumntype{L}[1]{>{\raggedright\let\newline\\\arraybackslash\hspace{0pt}}m{#1}}
\newcolumntype{C}[1]{>{\centering\let\newline\\\arraybackslash\hspace{0pt}}m{#1}}
\newcolumntype{R}[1]{>{\raggedleft\let\newline\\\arraybackslash\hspace{0pt}}m{#1}}

\def\OurAlgorithm{DreamFLEX}

\title{\LARGE \bf \OurAlgorithm: Learning Fault-Aware Quadrupedal Locomotion Controller for Anomaly Situation in Rough Terrains}

\author{Seunghyun Lee, I Made Aswin Nahrendra, Dongkyu Lee, Byeongho Yu, Minho Oh, and Hyun Myung$^{*}$\thanks{$^*$Corresponding author: Hyun Myung}
  \thanks{The authors are with the School of Electrical Engineering, KAIST (Korea Advanced Institute of Science and Technology), Daejeon, 34141, Republic of Korea. {\tt\scriptsize \{kevin9709, anahrendra, dklee, bhyu, minho.oh, hmyung\}@kaist.ac.kr}   	 
  }
}

\begin{document}
\maketitle
\thispagestyle{empty}
\pagestyle{empty}

\begin{abstract}
Recent advances in quadrupedal robots have demonstrated impressive agility and the ability to traverse diverse terrains.
However, hardware issues, such as motor overheating or joint locking, may occur during long-distance walking or traversing through rough terrains leading to locomotion failures.
Although several studies have proposed fault-tolerant control methods for quadrupedal robots, there are still challenges in traversing unstructured terrains.
In this paper, we propose \OurAlgorithm, a robust fault-tolerant locomotion controller that enables a quadrupedal robot to traverse complex environments even under joint failure conditions.
\OurAlgorithm\ integrates an explicit failure estimation and modulation network that jointly estimates the robot's joint fault vector and utilizes this information to adapt the locomotion pattern to faulty conditions in real-time, enabling quadrupedal robots to maintain stability and performance in rough terrains. 
Experimental results demonstrate that \OurAlgorithm\ outperforms existing methods in both simulation and real-world scenarios, effectively managing hardware failures while maintaining robust locomotion performance.

\end{abstract}

\section{Introduction}
\label{sec:intro}

Along with the development of autonomous navigation systems, various robot platforms capable of executing these systems have emerged. 
Among them, quadrupedal robots~\cite{anymalc, minicheetah, hound} have received significant interest from the research community owing to their high agility and ability to traverse diverse terrains.
Along with this growing interest, the practical applications of quadruped robots across various fields are also increasing~\cite{anymal_field, darpa, qrc}.

To maximize the agility of quadrupedal robots, a robust controller is essential for ensuring the robot adapts and maintains its stability under various conditions while avoiding hardware damage.
Recently, thanks to the development of deep reinforcement learning~(DRL), there have been large improvements in learning dynamic locomotion skills, such as robust walking in complex environments~\cite{rma, miki_at_el, fuzzy, dreamwaq, egocentric_locomotion, deformable, moral, dtc}, recovery from falls~\cite{recovery, mela, dreamriser}, and even more dynamic movements such as parkour~\cite{barkour, robotparkour, anymal_parkour, extreme_parkour}.
These studies show remarkable performances in traversing various terrains and overcoming challenging obstacles.
However, most of these studies assume normal conditions without taking into account potential hardware failures.

Despite the advancements in quadrupedal locomotion control, unexpected situations can arise during real-world experiments.
During long-distance walking or navigating challenging terrains, hardware failures such as motor overheating or joint locking, often caused by sudden external impact, can occur. These issues cause the robot to fall, significantly damaging the platform or the surrounding environment, requiring human intervention.

To address this problem, recent works have shown a fault-tolerant control of quadrupedal robots demonstrating its robustness.
Previous works based on meta-learning~\cite{famle, metaRL} and reinforcement learning~\cite{acdr} have shown promising results in simulations, leaving opportunities to extend their applicability in real-world.
Recent studies~\cite{saving_the_limping, FT_Kim, FT_Net} proposed a fault-tolerant control based on DRL, demonstrating their robustness via real-world experiments.
However, these studies had a limited demonstration for overcoming various rough terrains.
These environments, where the robot is particularly vulnerable to failure, present ongoing challenges for ensuring robust performance for fault-tolerant locomotion control.

\begin{figure}[t!]
	\captionsetup{font=footnotesize}
	\centering
\includegraphics[width=0.48\textwidth]{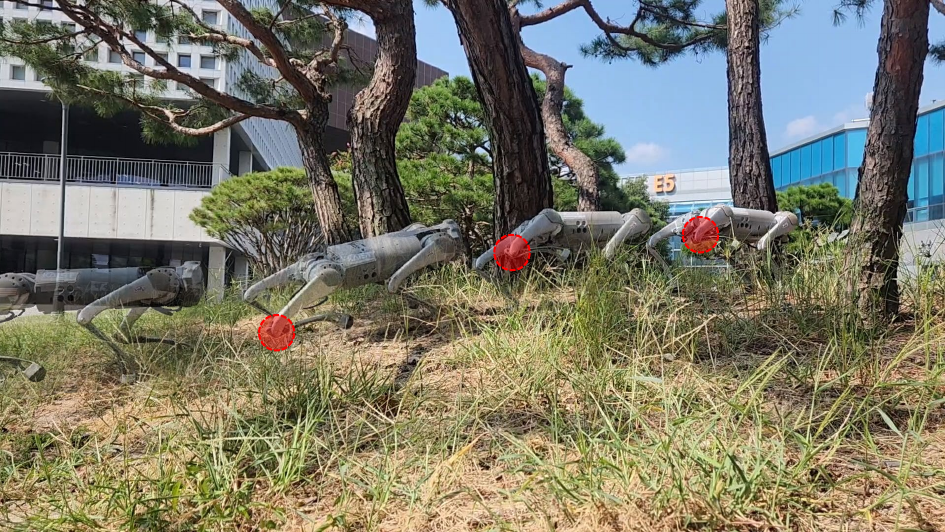}
\caption{\OurAlgorithm\ enables the robot to autonomously detect leg joint faults during locomotion~(with the rear right calf joint highlighted by a red circle). Subsequently, the policy is robustly adapted to overcome rough terrains, even under faulty conditions.}
	\label{fig:intro}
	\vsfig
\end{figure}

In this paper, we propose a robust fault-tolerant control framework called \textbf{\OurAlgorithm}, which stands for Dream walking for Fault-aware Locomotion control using EXplicit failure estimation and modulation, that can handle joint faults that may occur in quadrupedal robots during locomotion and enable robust locomotion through unstructured environments, as shown in Fig.~\ref{fig:intro}.
Our proposed framework is based on DreamWaQ~\cite{dreamwaq}, a state-of-the-art locomotion controller for quadrupedal robots, trained via DRL algorithm with only proprioceptive information.
\OurAlgorithm\ jointly infers the terrain properties which is essential for robust locomotion in various terrains, estimates the robot's state, whether it is in normal or faulty conditions, and failure points.
We leverage this information to adaptively change the locomotion pattern to maintain stable locomotion.
We validated the robustness of \OurAlgorithm\ through real-world experiments using a Unitree Go1~\cite{Unitree}.
In summary, the main contributions of this paper are as follows:

\begin{itemize}
\item 

We introduce \OurAlgorithm, a robust fault-tolerant quadrupedal locomotion control framework that can address various hardware faults and traverse unstructured environments via deep reinforcement learning.

\item 
We propose a failure estimation and modulation network~(FEMNet) to estimate the explicit joint state of the robot as a joint fault vector and leverage it to modulate the latent embedding for the policy network, which is used to adjust the locomotion pattern adaptively. The proposed FEMNet significantly improves the controller's command tracking performance.

\item 
The robustness of \OurAlgorithm\ has been demonstrated in real-world unstructured environments even under various fault conditions\footnote[1]{Project site: \texttt{https://dreamflex.github.io/}}. 

\end{itemize}

\section{Methodology of \OurAlgorithm}
\label{sec:method}

The overall architecture of \OurAlgorithm\ is shown in Fig.~\ref{fig:overall_arch}.
\OurAlgorithm\ is trained via deep reinforcement learning with only proprioceptive information based on an asymmetric actor-critic architecture (Section~\ref{subsec:methodA}).
To deal with abnormal situations for quadrupedal robots, each agents are trained with randomly assigned normal or faulty conditions (Section~\ref{subsec:methodB}). 
Subsequently, a failure estimation and modulation network (FEMNet) estimates the failure vector and leverages it to modulate the latent embedding for the policy network (Section~\ref{subsec:methodC}). 
This framework enables the robot to autonomously estimate its failure joints and leverage this information to traverse various terrains.
The details are described in the following subsections.
 \subsection{Training a Fault-tolerant Locomotion Controller}\label{subsec:methodA}

The environment is represented as a partially observable Markov decision process~(POMDP), where the agent has limited access to the full information of the environment.
It is defined by a tuple $(\mathcal{S}, \mathcal{O}, \mathcal{A}, d_0, p, r, \gamma)$, where $\mathcal{S}$ is the state space, $\mathcal{O}$ is the observation space, $\mathcal{A}$ is the action space, $d_0$ is the initial state distribution of the environment, $p\colon \mathcal{S} \times \mathcal{A} \times \mathcal{S} \rightarrow [0, 1]$ is the state transition probability, $r\colon \mathcal{S} \times \mathcal{A} \rightarrow \mathbb{R}$ is the reward function, and $\gamma\in [0, 1)$ is the discount factor.

\begin{figure}[t!]
	\captionsetup{font=footnotesize}
	\centering
\includegraphics[width=0.48\textwidth]{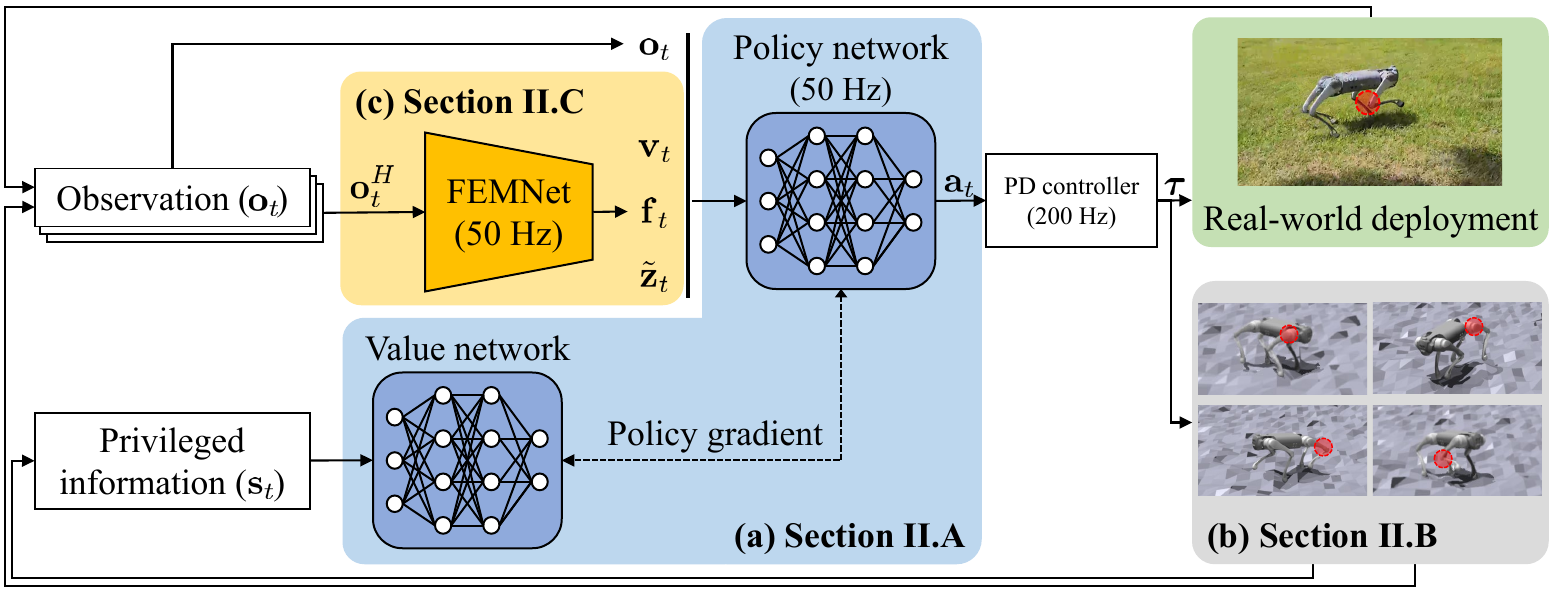}
    \caption{Overview of our proposed training framework, \OurAlgorithm. The framework consists of three main components:
	(a) learning a quadrupedal locomotion controller based on asymmetric actor-critic architecture, (b) randomly assigned joint failure scenarios, and (c) a failure estimation and modulation network (FEMNet) for estimating the failure vector, $\boldsymbol{\mathrm{f}}_t$, and modulating the latent embedding, $\tilde{\boldsymbol{\mathrm{z}}}_t$, for the policy network.
	}
	\label{fig:overall_arch}
	\vsfig
\end{figure}

Motivated by our previous study, DreamWaQ~\cite{dreamwaq}, \OurAlgorithm\ was designed based on an asymmetric actor-critic architecture~\cite{A2C}, which is efficient for implicit terrain property learning using only proprioceptive information.
The architecture, shown in Fig.~\ref{fig:overall_arch}(a), consists of two main networks: policy (actor) and value (critic) networks.
The policy network, $\pi_\phi(\boldsymbol{\mathrm{a}}_t|\boldsymbol{\mathrm{o}}_t,\boldsymbol{\mathrm{v}}_t, \boldsymbol{\mathrm{f}}_t, \boldsymbol{\mathrm{z}}_t)$ outputs the action $\boldsymbol{\mathrm{a}}_t\in\mathcal{A}$, given a partial observation $\boldsymbol{\mathrm{o}}_t\in\mathcal{O}$, body linear velocity $\boldsymbol{\mathrm{v}}_t$, joint fault vector $\boldsymbol{\mathrm{f}}_t$, and latent vector $\boldsymbol{\mathrm{z}}_t$. 
$\boldsymbol{\mathrm{f}}_t$ is a binary vector that indicates the state of the joints, which will be detailed in Section~\ref{subsec:methodB}.
During inference, $\boldsymbol{\mathrm{v}}_t$, $\boldsymbol{\mathrm{f}}_t$ and $\boldsymbol{\mathrm{z}}_t$ are estimated by a failure estimation and modulation network (FEMNet) and its details will be described in Section~\ref{subsec:methodC}. 
The value network, $V(\boldsymbol{\mathrm{s}}_t)$, is trained to evaluate the actions given the privileged state $\boldsymbol{\mathrm{s}}_t\in\mathcal{S}$. 
The proximal policy optimization (PPO) algorithm~\cite{PPO} is utilized to optimize the policy and value networks.

\subsubsection{Observation and Action Space}
The partial observation $\boldsymbol{\mathrm{o}}_t$ is the proprioceptive information obtained from the inertial measurement unit (IMU) and joint encoders defined as follows:
\begin{equation}
    \begin{gathered}
    \boldsymbol{\mathrm{o}}_t = [\boldsymbol{\mathrm{w}}_t\quad \boldsymbol{\mathrm{g}}_t\quad \boldsymbol{\mathrm{c}}_t\quad \boldsymbol{\mathrm{q}}_t\quad \dot{\boldsymbol{\mathrm{q}}}_t\quad \boldsymbol{\mathrm{a}}_{t-1}],
    \label{eq:observation}
    \end{gathered}
\end{equation}
where $\boldsymbol{\mathrm{w}}_t$, $\boldsymbol{\mathrm{g}}_t$, $\boldsymbol{\mathrm{c}}_t$, $\boldsymbol{\mathrm{q}}_t$, $\dot{\boldsymbol{\mathrm{q}}}_t$, and $\boldsymbol{\mathrm{a}}_{t-1}$ are the body angular velocity, gravity vector projected into the body frame, body velocity command, joint angle, joint angular velocity in the current time step $t$, and action values in the previous step, respectively.
The privileged information $\boldsymbol{\mathrm{s}}_t$ is the combination of the observation $\boldsymbol{\mathrm{o}}_t$ and the other environment information that are directly available in simulations as follows:
\begin{equation}
    \begin{gathered}
    \boldsymbol{\mathrm{s}}_t = [\boldsymbol{\mathrm{o}}_t\quad \boldsymbol{\mathrm{v}}_t\quad \boldsymbol{\mathrm{f}}_t\quad \boldsymbol{\mathrm{d}}_{t}\quad \boldsymbol{\mathrm{h}}_{t}], 
    \label{eq:previleged}
    \end{gathered}
\end{equation}
where $\boldsymbol{\mathrm{d}}_{t}$ denotes the external disturbance and $\boldsymbol{\mathrm{h}}_{t}$ denotes the local terrain height map around the robot.
The action $\boldsymbol{\mathrm{a}}_t$ denotes the joint angle command for each joint motor.
The desired joint angles of the robot are defined as $\boldsymbol{\mathrm{q}}_\text{des} = \boldsymbol{\mathrm{q}}_\text{def}\ +\ \boldsymbol{\mathrm{a}}_t$, where $\boldsymbol{\mathrm{q}}_\text{def}$ is the default joint angle when the robot is in a standing position.
$\boldsymbol{\mathrm{q}}_\text{des}$ is tracked by a proportional-derivative (PD) controller as $\boldsymbol{\mathrm{\tau}} = K_p(\boldsymbol{\mathrm{q}}_\text{des} - \boldsymbol{\mathrm{q}}) - K_d\boldsymbol{\dot{\mathrm{q}}}$, 
where $K_p$ and $K_d$ are the proportional and derivative gains, respectively, and $\boldsymbol{\mathrm{\tau}}$ is the torque command for each joint motor.

\subsubsection{Rewards}
Most DRL-based locomotion controllers utilize two types of reward functions: task and style rewards.
Task rewards are designed to track the linear and angular velocity commands of the robot and style rewards are designed to maintain stability and minimize unnecessary motions.

However, we discovered that the existing reward functions, designed without considering faulty situations, are insufficient to produce fault-tolerant controllers with natural locomotion gaits. 
Therefore, beyond the basic rewards from our previous work~\cite{dreamwaq}, we introduce additional fault-tolerant rewards to ensure the robot exhibits natural movements even in failure scenarios, as shown in Table~\ref{table:reward}.
There are two main objectives of fault-tolerant rewards: ensure stable locomotion even in fault scenarios using the non-faulty legs (i.e., by considering feet air time, foot clearance, and Raibert) and minimize the unnecessary movement and ground contact of the faulty leg to prevent it from hindering locomotion. (i.e., by considering faulty leg joint motion and faulty leg contact).
The total reward is calculated as $r_t(\boldsymbol{\mathrm{s}}_t, \boldsymbol{\mathrm{a}}_t) = \sum_i r_iw_i$ at each time step, where $r_i$ and $w_i$ are the $i$-th reward function and its weight, respectively.

\begin{table}[t!]
	\centering
	\captionsetup{font=footnotesize}
	\renewcommand{\arraystretch}{1.25}
	\caption{Fault-tolerant reward functions $r_i$ and their weights $w_i$. 
$(\cdot)^\text{des}$ and $(\cdot)^\text{thr}$ denote the desired and threshold values, respectively. 
	$(\cdot)^n$ and $(\cdot)^f$ denote normal and faulty legs, respectively.
	$\boldsymbol{\mathrm{t}}_{\text{air, foot}}$, $\boldsymbol{\mathrm{p}}_\text{foot}$, and $\boldsymbol{\mathrm{f}}_\text{foot}$ are the air time of the foot, the foot position \text{w.r.t.} the body frame and the foot contact force, respectively.}
	{\scriptsize		
		\begin{tabular}{l|c|c}
			\toprule\midrule
			$\textbf{Fault-tolerant reward}$ & \begin{tabular}[c]{@{}c@{}}Equation ($r_i$)\end{tabular} & \begin{tabular}[c]{@{}c@{}}Weight ($w_i$)\end{tabular} \\ \midrule
			Feet air time          &   $\sum(\boldsymbol{\mathrm{t}}^{n}_{\text{air, foot}} - \boldsymbol{\mathrm{t}}^{n, \text{des}}_{\text{air, foot}})$          
			&            $1.5$           \\
			Foot clearance                  &   $(\boldsymbol{\mathrm{p}}^{n}_{z, \text{foot}} - \boldsymbol{\mathrm{p}}^{n, \text{des}}_{z, \text{foot}})^2\cdot \boldsymbol{\mathrm{v}}_{xy, \text{foot}}$              
			&            $-0.5$           \\
			Raibert                      &   $(\boldsymbol{\mathrm{p}}^{n}_{xy, \text{foot}} - \boldsymbol{\mathrm{p}}^{n, \text{des}}_{xy, \text{foot}})^2$            
			&            $-1 \times 10^{-5}$           \\
			Faulty leg joint motion                  &   $(\boldsymbol{\mathrm{q}}^{f} - \boldsymbol{\mathrm{q}}^{f, \text{des}})^2$             
			&            $-0.2$           \\
			Faulty leg contact                &  $\mathbf{1}_{\boldsymbol{\mathrm{f}}^{f}_{\text{foot}} > \boldsymbol{\mathrm{f}}^{f, \text{thr}}_{\text{foot}}}$            &            $-0.1$    \\ 
			\midrule\bottomrule
		\end{tabular}
	}
	\label{table:reward}
	\vstab
\end{table} 
\subsubsection{Training Details}\label{subsubsec:training}
We utilize a terrain curriculum~\cite{massiveRL} to progressively learn walking policies from easy to difficult terrains including smooth, rough, discretized, and stair terrains, ensuring robust locomotion even on rough terrain.
Moreover, unlike normal quadrupedal locomotion, a significant issue arises when learning to walk with three legs in failure scenarios. 
The robot tends to converge on a walking pattern where the legs do not lift high enough, resulting in a sliding motion on the ground and this insufficient foot clearance can cause a large sim-to-real gap.
To address this issue, additional fractal noise terrains~\cite{whole_body_control} are added during training progress with ten levels of height amplitude within $[0.04, 0.12]~\text{m}$, encouraging the robot to lift its legs higher and achieve a more reliable walking pattern.

 \subsection{Joint Fault}\label{subsec:methodB}
\begin{figure}[t!]
	\captionsetup{font=footnotesize}
	\centering
	\includegraphics[width=0.48\textwidth]{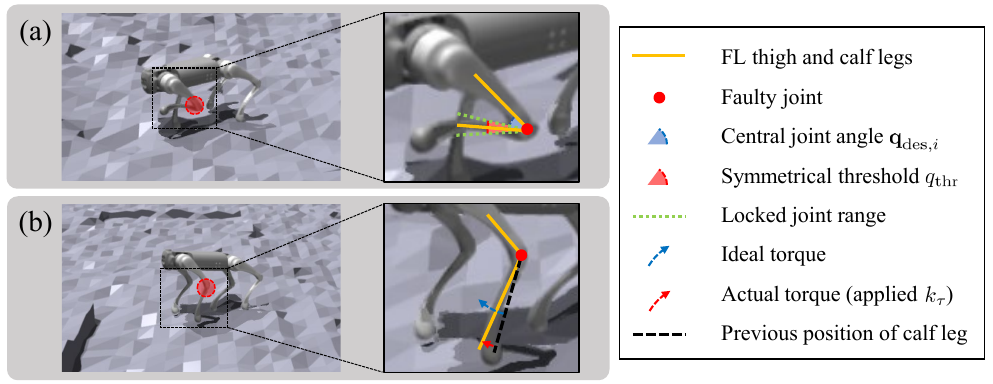}
    \caption{Examples of the fault scenarios in the front left calf joint of the quadrupedal robot: (a) locked joint and (b) weakened motor.}
	\label{fig:joint_fault}
	\vsfig
\end{figure}

Quadrupedal robots typically encounter two major unexpected failure situations: 1)~locked joint and 2)~weakened motor.
To address these hardware issues that can occur in real-world, each agent is randomly assigned with either normal or faulty conditions in the training process as shown in Fig.~\ref{fig:joint_fault} and given with the corresponding joint fault vector.
The details are described in the followings.

\subsubsection{Locked Joint}\label{subsec:locked_joint} 
The locked joint is a situation when the robot's joint motor becomes immobilized due to a significant impact, usually from a collision, preventing the joint from moving.
We implement the locked joint scenarios by applying a clipping method to the desired joint angle calculated from the action as follows:
\begin{equation}
    \begin{gathered}
        \boldsymbol{\mathrm{q}}_{\text{des}, i} \leftarrow
        \begin{cases}
            \boldsymbol{\mathrm{q}}_{\text{des}, i}, & \mbox{if }i\mbox{-th joint is normal,} \\
            \text{Clip}(\boldsymbol{\mathrm{q}}_{\text{des}, i}, & \mbox{otherwise,} \\ \qquad\, \boldsymbol{\mathrm{q}}_{\text{cen}, i}\pm q_{\text{thr}}),
        \end{cases}
    \label{eq:locked_joint}
    \end{gathered}
\end{equation}
where $\boldsymbol{\mathrm{q}}_{\text{cen}}$ is the central joint angle of allowed joint range and $q_{\text{thr}}$ is the symmetrical threshold value (e.g. $q_{\text{thr}}=0.05~\text{rad}$). 

\subsubsection {Weakened Motor}\label{subsec:weakened_motor}
The motor is regarded as weakened when the joint motor fails to function correctly due to insufficient power supply or overheating from prolonged use, causing inadequate electrical signals to be delivered to the motor.
To simulate a weakened motor, we introduce the torque efficiency factor $k_{\tau}$ to reduce the actual torque command of the faulty joint as follows:
\begin{equation}
    \begin{gathered}
        \boldsymbol{\mathrm{\tau}}_i =
        \begin{cases}
            \boldsymbol{\mathrm{\tau}}_i, & \mbox{if }i\mbox{-th joint is normal,} \\
            k_{\tau} \cdot \boldsymbol{\mathrm{\tau}}_i, & \mbox{otherwise.}
        \end{cases}
    \label{eq:weakened_motor}
    \end{gathered}
\end{equation}

\subsubsection {Training With Joint Fault}\label{subsec:training_joint_failure}
During the training process, each agent is randomly assigned with either normal or faulty conditions, and the fault joint is uniformly sampled from all joints of quadrupedal robots.
To simplify the explicit representation of each joint motor's state, we introduce the joint fault vector $\boldsymbol{\mathrm{f}}_t\in\mathbb{R}^n$, a binary vector corresponding to the $n$ joints (usually, $n=12$) of a quadrupedal robot in the order of the front left (FL), front right (FR), rear left (RL), and rear right (RR) legs with each hip, thigh, and calf joints.
$\boldsymbol{\mathrm{f}}_t$ is defined as follows:
\begin{equation}
    \begin{gathered}
        \boldsymbol{\mathrm{f}}_{t, i} =
        \begin{cases}
            0, & \mbox{if }i\mbox{-th joint is normal,} \\
            1, & \mbox{otherwise.}
        \end{cases}
    \label{eq:joint_fault_vector}
    \end{gathered}
\end{equation}

Moreover, $\boldsymbol{\mathrm{q}}_{\text{cen}}$ and $k_{\tau}$ in (\ref{eq:locked_joint}) and (\ref{eq:weakened_motor}) are also randomly sampled from the uniform distribution as follows: $\boldsymbol{\mathrm{q}}_{\text{cen}}\sim\mathcal{U}(q_L, q_U)$ and $k_{\tau}\sim\mathcal{U}(k_L,0.25)$, where $q_L$, $q_U$, and $k_L$ are the lower, upper bounds of the joint angle, and the lower bound of the torque efficiency, respectively.
However, when trained with a large range initially, the training process might be unstable due to large variance and sometimes fails to converge.
Therefore, we introduce a failure curriculum mechanism to gradually increase the sampling range of the failure parameters, inspired by~\cite{acdr, rapid}.
When the episodic sum of the task reward $r_\text{task}$ exceeds the success threshold value $r_{\text{thr}}$, the curriculum failure mechanism adjusts the range of $\boldsymbol{\mathrm{q}}_{\text{cen}}$ and $k_{\tau}$ as follows:
\begin{equation}
    \begin{aligned}
        q_L &\leftarrow \text{max}(q_L-\delta_q, q_{\text{min}}), \\
        q_U &\leftarrow \text{min}(q_U+\delta_q, q_{\text{max}}), \\
        k_L &\leftarrow \text{max}(k_L-\delta_k, 0),
    \end{aligned}
    \label{eq:joint_fault_vector}
\end{equation}
where $q_{\text{min}}$ and $q_{\text{max}}$ are the actual lower and upper bounds of the joint angle, respectively. $\delta_q$ and $\delta_k$ are the step sizes for adjusting the range of $\boldsymbol{\mathrm{q}}_{\text{cen}}$ and $k_{\tau}$, respectively. 
We set $\delta_q = 0.05~\text{rad}$ and $\delta_k = 0.0125$.

 \subsection{Failure Estimation and Modulation Network}\label{subsec:methodC}
As described in Section~\ref{subsec:methodA}, the policy network requires $\boldsymbol{\mathrm{v}}_t$, $\boldsymbol{\mathrm{f}}_t$, and $\boldsymbol{\mathrm{z}}_t$ as input, and these values are not directly observable from proprioceptions.
% Prior study estimates explicit state of the robot using only proprioception~\cite{concurrent} and improves the robustness of the locomotion policy via CENet~\cite{dreamwaq} that jointly estimates $\boldsymbol{\mathrm{v}}_t$ and $\boldsymbol{\mathrm{z}}_t$ using a $\beta$-variational auto-encoder~($\beta$-VAE)~\cite{beta_vae}.
Prior study~\cite{concurrent} estimates explicit state of the robot using only proprioception and DreamWaQ~\cite{dreamwaq} improves the robustness of the locomotion policy via CENet that jointly estimates $\boldsymbol{\mathrm{v}}_t$ and $\boldsymbol{\mathrm{z}}_t$ using a $\beta$-variational auto-encoder~($\beta$-VAE)~\cite{beta_vae}.
Additionally, prior works implicitly infer the robot's fault states using a learned estimator network~\cite{FT_Kim,FT_Net}.

Quadrupedal locomotion patterns must be changed depending on their joint conditions and adapted based on the specific location of the fault to achieve stable locomotion across various scenarios. 
In other words, the latent vector, which implicitly infers implicit information about the robot and the surrounding environment and significantly impacts walking performance, needs to adequately incorporate the information from the joint fault vector. 

By extending those ideas, we propose a failure estimation and modulation network~(\textbf{FEMNet}) that jointly learns to estimate the joint fault vector and latent vector.  
Subsequently, we leverage the joint fault vector to modulate the latent vector, as shown in Fig.~\ref{fig:estimation_modulation}.
FEMNet is composed of two main components as follows:

\subsubsection{Failure Estimation Model}\label{subsec:est}
A failure estimation model is designed based on the CENet~\cite{dreamwaq} and is modified to estimate the joint fault vector $\boldsymbol{\mathrm{f}}_t$, as shown in Fig.~\ref{fig:estimation_modulation}(a).
The encoder network of FEMNet takes the partial observation history $\boldsymbol{\mathrm{o}}^H_t$ as input ($\text{i.e.,}~[\boldsymbol{\mathrm{o}}_t~\boldsymbol{\mathrm{o}}_{t-1}~\cdots~\boldsymbol{\mathrm{o}}_{t-N}]^T$, $N=5$) and outputs $\boldsymbol{\mathrm{v}}_t$, $\boldsymbol{\mathrm{f}}_t$, and $\boldsymbol{\mathrm{z}}_t$. 
The decoder network takes $\boldsymbol{\mathrm{z}}_t$ as input and reconstructs the next observation $\boldsymbol{\mathrm{o}}_{t+1}$ to induce consistent state transition in the latent space.
The loss function $\mathcal{L}_{\text{FEM}}$ is defined as follows:

\begin{equation}
    \begin{gathered}
        \mathcal{L}_{\text{FEM}} = \mathcal{L}_{\text{est,f}} + \mathcal{L}_{\text{est,v}} + \mathcal{L}_{\text{VAE}},
    \label{eq:loss_cenet}
    \end{gathered}
\end{equation}
where $\mathcal{L}_{\text{est,f}}$ and $\mathcal{L}_{\text{est,v}}$ are the explicit estimation losses of the joint fault vector and linear velocity, respectively, and $\mathcal{L}_{\text{VAE}}$ is the VAE loss.
For joint fault vector estimation, we used a binary cross entropy (BCE) loss between the ground truth $\bar{\boldsymbol{\mathrm{f}}}_t$ and the estimated values of the joint fault vector $\boldsymbol{\mathrm{f}}_t$.
BCE loss directly optimizes the predicted probabilities for binary estimation, which yields better accuracy and performance in binary tasks than MSE loss. 
Therefore, BCE loss is utilized for the joint fault vector estimation, defined as follows: 
\begin{equation}
    \begin{gathered}
        \mathcal{L}_{\text{est,f}} = BCE(\boldsymbol{\mathrm{f}}_t, \bar{\boldsymbol{\mathrm{f}}}_t).
    \label{eq:est_loss}
    \end{gathered}
\end{equation}
Moreover, motivated by CENet~\cite{dreamwaq}, we designed $\mathcal{L}_{\text{est,v}}$ as a mean squared error (MSE) loss between the ground truth $\bar{\boldsymbol{\mathrm{v}}}_t$ and the estimated values of linear velocity $\boldsymbol{\mathrm{v}}_t$ and $\mathcal{L}_{\text{VAE}}$ as a combination of reconstruction loss and latent loss utilizing MSE and Kullback-Leibler (KL) divergence~\cite{KL}, respectively.

\subsubsection{Modulation Model}\label{subsec:mod}

\begin{figure}[t!]
	\captionsetup{font=footnotesize}
	\centering
    \includegraphics[width=0.48\textwidth]{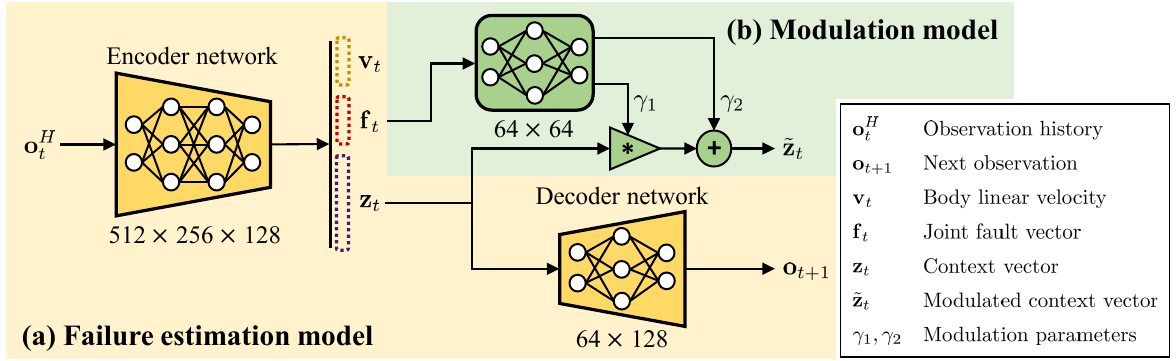}
	\caption{The architecture of FEMNet. (a) FEMNet jointly estimates the body linear velocity $\boldsymbol{\mathrm{v}}_t$, joint fault vector $\boldsymbol{\mathrm{f}}_t$, and latent vector $\boldsymbol{\mathrm{z}}_t$, and (b) modulates the latent vector based on the joint fault vector.}
	\label{fig:estimation_modulation}
	\vsfig
\end{figure}

Motivated by prior works~\cite{film,user_conditioned}, we propose a modulation model that enables the robot to adopt appropriate gait behaviors based on its current condition. 
In other words, the modulation model conditions the latent vector $\boldsymbol{\mathrm{z}}_t$ based on the joint fault vector $\boldsymbol{\mathrm{f}}_t$ and its architecture is shown in Fig.~\ref{fig:estimation_modulation}(b).
The modulation layer receives $\boldsymbol{\mathrm{f}}_t$ as an input and 
learns modulation parameters $\gamma_1$ and $\gamma_2$.
Subsequently, the latent vector $\boldsymbol{\mathrm{z}}_t$ is passed through the affine transformation with the learned parameters $\gamma_1$ and $\gamma_2$.
The formulation of the affine transformation is as follows:
\begin{equation}
    \begin{gathered}
        \tilde{\boldsymbol{\mathrm{z}}}_t = \gamma_1\cdot \boldsymbol{\mathrm{z}}_t +\gamma_2,
    \label{eq:film}
    \end{gathered}
\end{equation}
where $\tilde{\boldsymbol{\mathrm{z}}}_t$ is the modulated latent vector.
After the transformation, $\tilde{\boldsymbol{\mathrm{z}}}_t$ is used as an input to the policy network, allowing the policy to leverage the modulated latent vector to adjust the locomotion pattern based on the fault information adaptively.

\section{Experiments}
\label{sec:exp}

\subsection{Training in Simulation}\label{subsec:training}
The training was conducted with 4,096 agents in parallel on the Isaac Gym simulator~\cite{massiveRL, isaac_gym} using the Unitree Go1 model~\cite{Unitree}. 
For DRL training, we used PPO algorithm with the hyperparameters shown in Table~\ref{table:training_parameter}.
All networks were trained on a NVIDIA A5000 GPU and full training using our proposed algorithm took approximately two hours of wall-clock time to complete.
To evaluate the performance of our proposed method, we conducted a comparative analysis:
\begin{itemize}
\item \textbf{Kim \textit{et al.}}~\cite{FT_Kim}: The policy was trained with reward functions and implicit joint state estimator proposed in~\cite{FT_Kim} using the teacher-student training framework. 

\item \textbf{FT-Net}~\cite{FT_Net}: 
The policy was trained with 1D convolutional neural network~(CNN)-based adaptor and reward functions proposed in~\cite{FT_Net}.

\item \textbf{\OurAlgorithm\ w/o joint fault vector}: Without explicit joint fault vector $\boldsymbol{\mathrm{f}}_t$ explained in Section~\ref{subsec:methodB}. 
That is, the policy network consists of a FEMNet without the joint fault vector estimation and modulation model, same as the previous work~\cite{dreamwaq}. 

\item \textbf{\OurAlgorithm\ w/o modulation}: Without the modulation model explained in Section~\ref{subsec:mod}. The joint fault vector $\boldsymbol{\mathrm{f}}_t$ is only used as a direct input to the policy network.

\item \textbf{\OurAlgorithm}: Our proposed method.
\end{itemize}
Each method described above was trained within the same simulation environment, utilizing a uniform curriculum strategy and failure mechanism. 
We used the same hyperparameters for training and all the methods above were trained using only proprioception.

\begin{figure}[t!]
	\captionsetup{font=footnotesize}
	\centering
\includegraphics[width=0.48\textwidth]{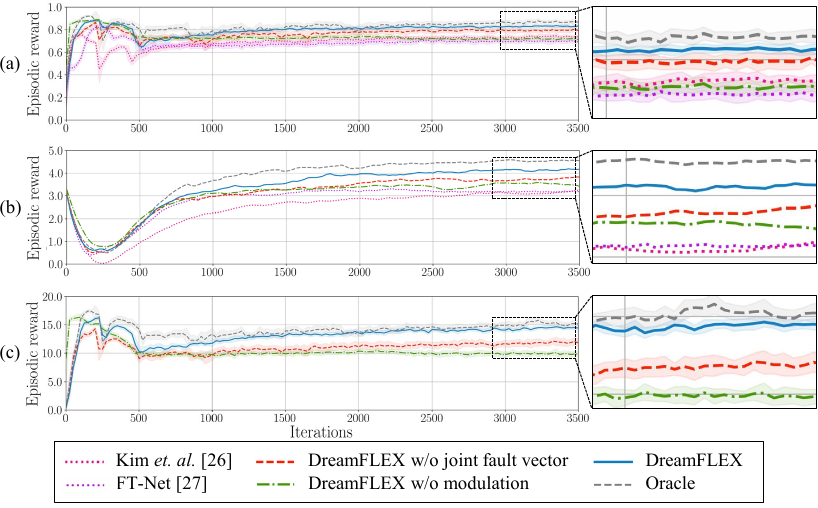}
\caption{Learning curves of the (a) linear velocity tracking, (b) terrain level, and (c) total rewards. The higher the value, the better the performance. As an upper bound performance, the oracle policy was trained by directly accessing the privileged information. Except for the oracle policy, \OurAlgorithm\, the blue solid line, outperforms all the compared algorithms.
}
	\label{fig:learning_curve}
	\vsfig
\end{figure}

The learning curves of \OurAlgorithm\ and the compared algorithms in simulation are shown in Fig.~\ref{fig:learning_curve}.
The learning curves were evaluated based on the linear velocity tracking, terrain level (i.e., the difficulty of the terrain the robot can traverse), and total rewards.
Kim \textit{et al.}~\cite{FT_Kim} and FT-Net~\cite{FT_Net} are excluded from the total reward graph because their reward functions differ from those proposed in this study.
Except for the oracle policy which can directly access the privileged information, \OurAlgorithm\ outperforms all the compared algorithms in terms of linear velocity tracking and terrain level.
This result indicates that \OurAlgorithm\ can effectively overcome various terrains and track the command velocity even in fault conditions.
Moreover, the proposed $\boldsymbol{\mathrm{f}}_t$ and modulation module show a significant performance gain in the total reward values, further confirming that the proposed method is the most effective for learning more robust locomotion even in fault conditions. 

\begin{table}[t!]
	\centering
	\captionsetup{font=footnotesize}
	\renewcommand{\arraystretch}{1.25}
	\caption{Training hyperparameters.}
	{\scriptsize		
		\begin{tabular}{lc|lc}
\toprule\midrule
PPO clip range 			& 0.2 		& Discount factor		& 0.99\\
			GAE factor 				& 0.95 		& Minibatches per epoch & 4\\
			Learning rate 			& $10^{-3}$ & Optimizer 			& Adam~\cite{adam} \\
			Activation function		& ELU~\cite{elu}		\\
\midrule\bottomrule
		\end{tabular}
	}
	\label{table:training_parameter}
	\vstab
\end{table}  \subsection{Simulated Experiments}\label{subsec:sim_exp}
To evaluate the performance of \OurAlgorithm, we conducted simulated experiments on a desktop PC with Intel Core i7-13700F CPU, 32GB RAM, and NVIDIA RTX 3080 GPU.
During experiments, we assigned joint fault to robots with fixed parameters: $\boldsymbol{\mathrm{q}}_\text{cen} = \boldsymbol{\mathrm{q}}_\text{def}$ (locked joint) and $k_{\tau} = 0.0$ (weakened motor).
\subsubsection{Command Tracking Performance}
\begin{table*}[t!]
    \centering
    \renewcommand{\arraystretch}{1.25}  \setlength{\tabcolsep}{5.1pt}
    \captionsetup{font=footnotesize}
    \caption{Absolute tracking error (ATE) for following the given linear velocity. 
    The best and second best results are indicated in \textbf{bold} and \underline{underline}, respectively. (Units:~$\text{m/s}$)}
    {\scriptsize
    \begin{tabular}{l|c|c|c|c|c|c|c|c|c|c|c|c!{\vrule width 1.2pt}c}
		\midrule\bottomrule
        & \multicolumn{6}{c|}{Locked Joint} & \multicolumn{6}{c!{\vrule width 1.2pt}}{Weakened Motor} \\ \cline{2-13} & \multicolumn{3}{c|}{Front} & \multicolumn{3}{c|}{Rear} & \multicolumn{3}{c|}{Front} & \multicolumn{3}{c!{\vrule width 1.2pt}}{Rear} & Mean \\ \cline{2-13} & Hip & Thigh & Calf & Hip & Thigh & Calf & Hip & Thigh & Calf & Hip & Thigh & Calf \\ \toprule 

        Kim \textit{et al.}~\cite{FT_Kim} & 0.3143 & 0.3965 & 0.4062 & 0.3341 & 0.3169 & 0.3921 & 0.3952 & 0.4052 & 0.3165 & 0.4599 & 0.3789 & 0.3356 & 0.3710 \\ 

        FT-Net~\cite{FT_Net} & 0.2634 & 0.3300 & 0.2701 & 0.2694 & 0.2988 & 0.3184 & 0.2884 & 0.2593 & 0.2738 & 0.2915 & 0.2625 & 0.2916 & 0.2848 \\ 

        \makecell[l]{ \OurAlgorithm\ w/o joint fault vector} & 0.2745 & 0.2981 & \underline{0.2068} & \underline{0.2488} & \underline{0.2516} & 0.4070 & 0.2750 & 0.2992 & \underline{0.2078} & 0.2480 & \underline{0.2510} & 0.4065 & 0.2812\\ 

        \makecell[l]{ \OurAlgorithm\ w/o modulation} & \underline{0.2082} & \underline{0.2620} & 0.2397 & 0.2533 & 0.2698 & \underline{0.2755} & \underline{0.2493} & \underline{0.2511} & 0.2304 & \underline{0.2074} & 0.2737 & \textbf{0.1904} & \underline{0.2426}\\ 

        \textbf{\OurAlgorithm} & \textbf{0.1581} & \textbf{0.1761} & \textbf{0.1587} & \textbf{0.1300} & \textbf{0.1539} & \textbf{0.1407} & \textbf{0.2032} & \textbf{0.1783} & \textbf{0.2021} & \textbf{0.1832} & \textbf{0.2419} & \underline{0.2172} & \textbf{0.1786}\\ \midrule\bottomrule
    \end{tabular}
    }
    \label{table:issacgym_trackingerror}
\end{table*} We evaluated the command tracking performance in the Isaac Gym simulator.
We tested 100 agents in parallel starting from the same initial position with same fault scenarios and commanded a forward velocity of 1.0~$\text{m/s}$ for 10 seconds.
In this experiment, we compared the performance on fractal noise terrain described in Section~\ref{subsubsec:training} with a max height amplitude of 0.12~$\text{m}$.
We utilized absolute tracking error (ATE) as the performance metric. The result is summarized in Table~\ref{table:issacgym_trackingerror}.

\OurAlgorithm\ consistently outperforms the other methods regardless of the failure scenarios.
Especially, the proposed FEMNet improves the tracking performance up to $36.49\%$ on average compared with the \OurAlgorithm\ without FEMNet~(i.e., w/o joint fault vector), thanks to its explicit joint state estimation and latent conditioning strategies that quickly adapted the policy's behavior. 
This result supports our claim that \OurAlgorithm\ can effectively adapt to various terrains and fault conditions, maintaining stable locomotion.

\subsubsection{Transition from Normal to Fault Conditions} 
\begin{figure}[t!]
	\captionsetup{font=footnotesize}
	\centering
\includegraphics[width=0.48\textwidth]{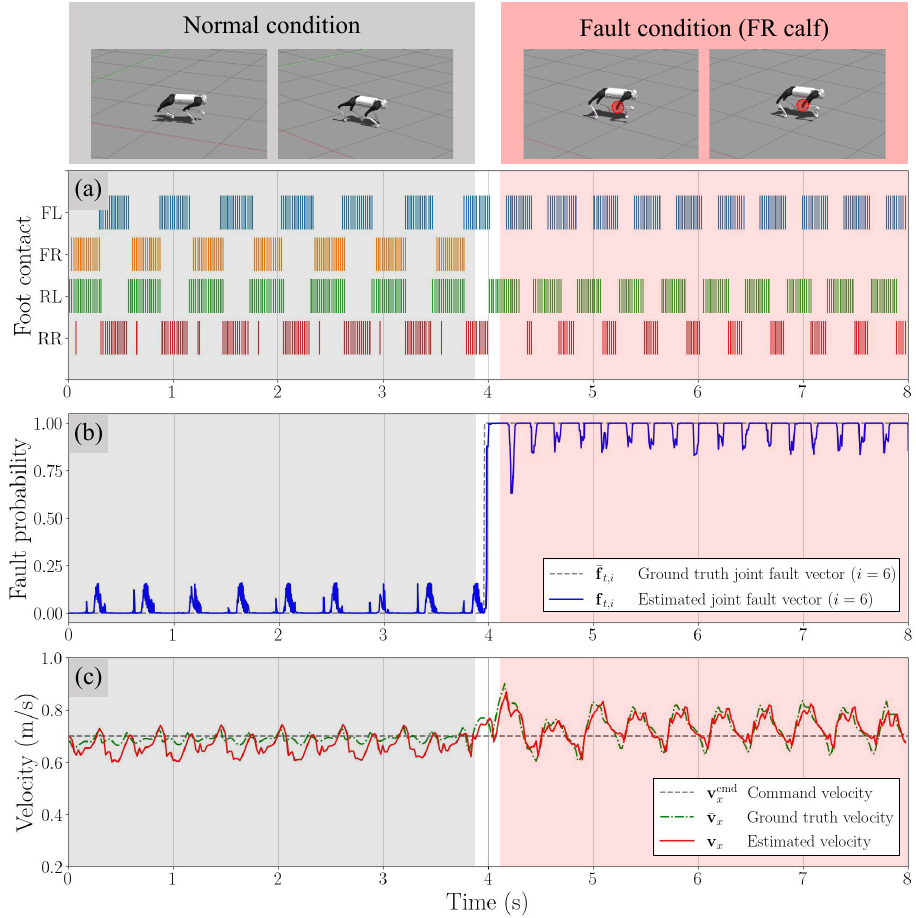}
\caption{
		Quantitative results of the transition from normal to fault conditions (locked FR calf joint) in the Gazebo simulator. Each graph shows the results of (a)~foot contact, (b)~joint fault vector estimation,
		and (c)~velocity command tracking and estimation performance over 8 seconds.
	}
	\label{fig:gazebo}
	\vsfig
\end{figure}
We conducted a simulation experiment in the Gazebo simulator~\cite{gazebo} to observe the behavior of the robot when the fault occurs.
We provided a forward velocity command 0.7~$\text{m/s}$ and locked the FR calf joint of the robot.
We observed the velocity tracking, estimation performance, and how the locomotion pattern and joint fault vector changed when a fault occurred, as shown in Fig.~\ref{fig:gazebo}.

Before the fault occurs, the robot walks normally with a trotting gait and $\boldsymbol{\mathrm{f}}_{t, i}~(i=6)$ is maintained at a low value which indicates that there is no fault in FR calf joint. 
However, when the fault occurs, the robot indicates failure points through increased $\boldsymbol{\mathrm{f}}_{t, i}$ and smoothly transitions its gait pattern, minimizing the use of the faulty leg and attemping to walk with the remaining legs.
Moreover, although there was a slight increase in velocity command tracking error right after the fault occurred, 
our proposed system shows small errors in both command tracking and velocity estimation, as shown in Fig.~\ref{fig:gazebo}(c). 
These results demonstrate that FEMNet can accurately estimate the linear velocity and joint fault vector even under sudden faults in real-time and effectively leverage it to adapt the locomotion pattern to maintain stable locomotion even under fault conditions. 

 \subsection{Real-world Experiments}\label{subsec:real_exp}
To demonstrate the practicality of \OurAlgorithm, various experiments were conducted using a Unitree Go1 quadruped robot in real-world unstructured environments.
The learned FEMNet and policy were deployed on the robot without additional fine-tuning.
During the real-world experiments, the control policy and FEMNet were executed at 50~Hz on an embedded NVIDIA Jetson Xavier NX.
We utilized PD controller as the low-level controller to track the desired joint angles, operating at 200~Hz with $K_p$ = 28.5 and $K_d$ = 0.72.
The computed torque commands were sent to the motor controller of the robot and joint faults are applied to the robots with fixed parameters as mentioned in Section~\ref{subsec:sim_exp}.
The experimental videos are available at the project site$^{1}$.

\begin{figure}[t]
	\captionsetup{font=footnotesize}
	\centering
	\includegraphics[width=0.48\textwidth]{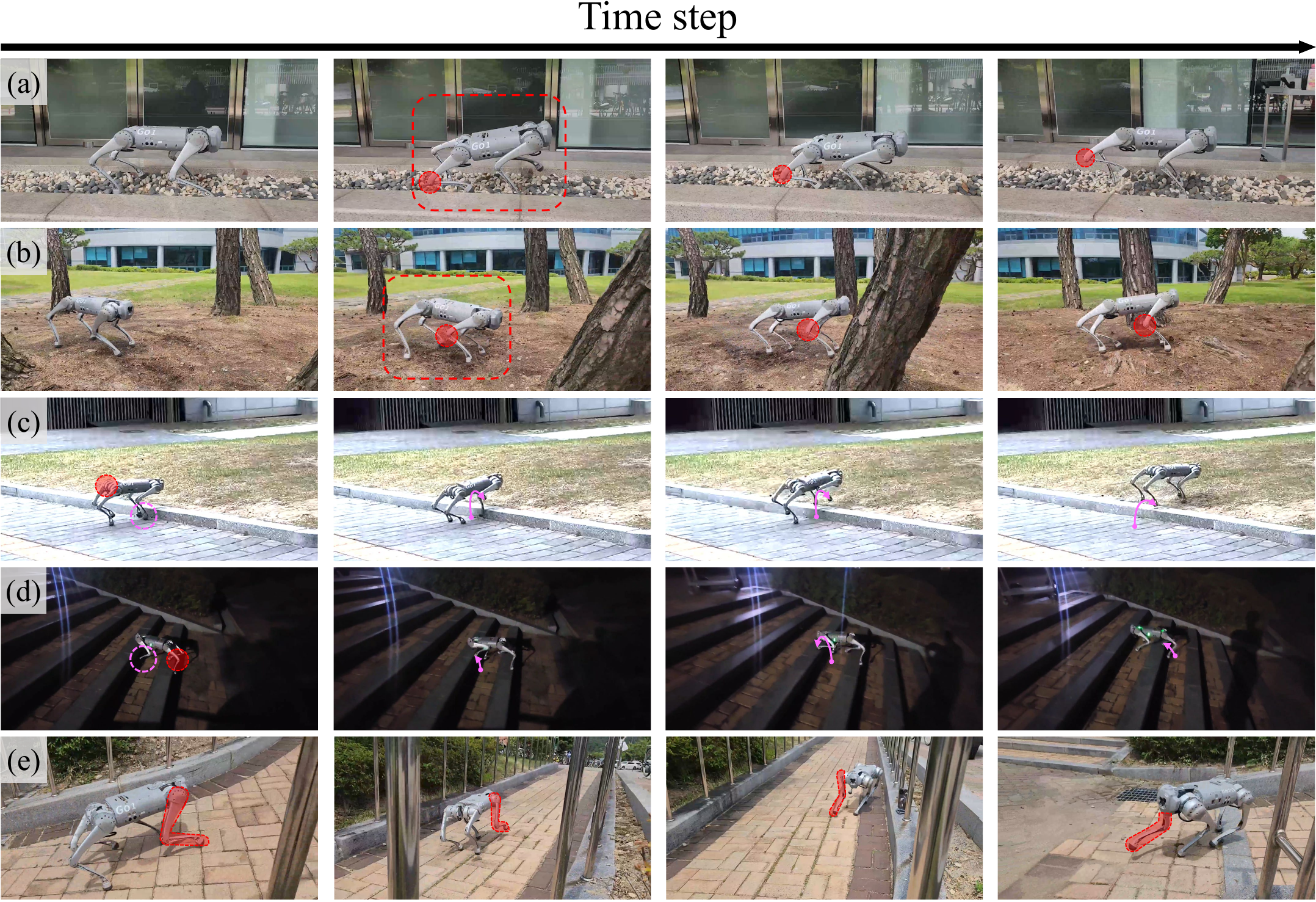}
	\caption{Our controller successfully overcame numerous unstructured terrains with various fault situations in real-world: 
(a) rough gravel,
	(b) slippery tree root,
	(c) discretized step,
	(d) upstairs, and
	(e) slope terrains.
	The red dotted circle and shape indicate the fault joint and the overheated FR leg, respectively.
	The pink circle and arrow in (c), (d) indicate the collision with the obstacles and its foot swing adaptation to overcome, respectively. 
	}
	\label{fig:outdoor_exp}
	\vsfig
\end{figure}

The tested outdoor environments include rough, slippery, and discretized terrains.
The robot controlled with \OurAlgorithm's policy successfully adapted its gaits to the environment and overcame the obstacles despite faults in the front and rear leg joints, as shown in Fig.~\ref{fig:outdoor_exp}.
The most challenging terrain was the step and stair terrains (Fig.~\ref{fig:outdoor_exp}(c),~(d)), which are quite challenging for small quadrupedal robots even in normal situations.
Thanks to the robustness of the policy trained with \OurAlgorithm, the robot, even with fault joint, could safely walk through the steps and stairs of 10~cm and 15~cm high, respectively.
Moreover, because the experiments were conducted during summer, the FR motors of the robot were heated up and became disabled~(Fig.~\ref{fig:outdoor_exp}(e)).
However, \OurAlgorithm\ quickly detected the fault location in real-time and adapted the gait pattern to maintain stable locomotion. 
This result demonstrates that the proposed algorithm works effectively in the actual failure of the robot, not only in manually-triggered failure scenarios as in Section~\ref{subsec:methodB}.

\begin{figure}[t!]
	\captionsetup{font=footnotesize}
	\centering
	\includegraphics[width=0.48\textwidth]{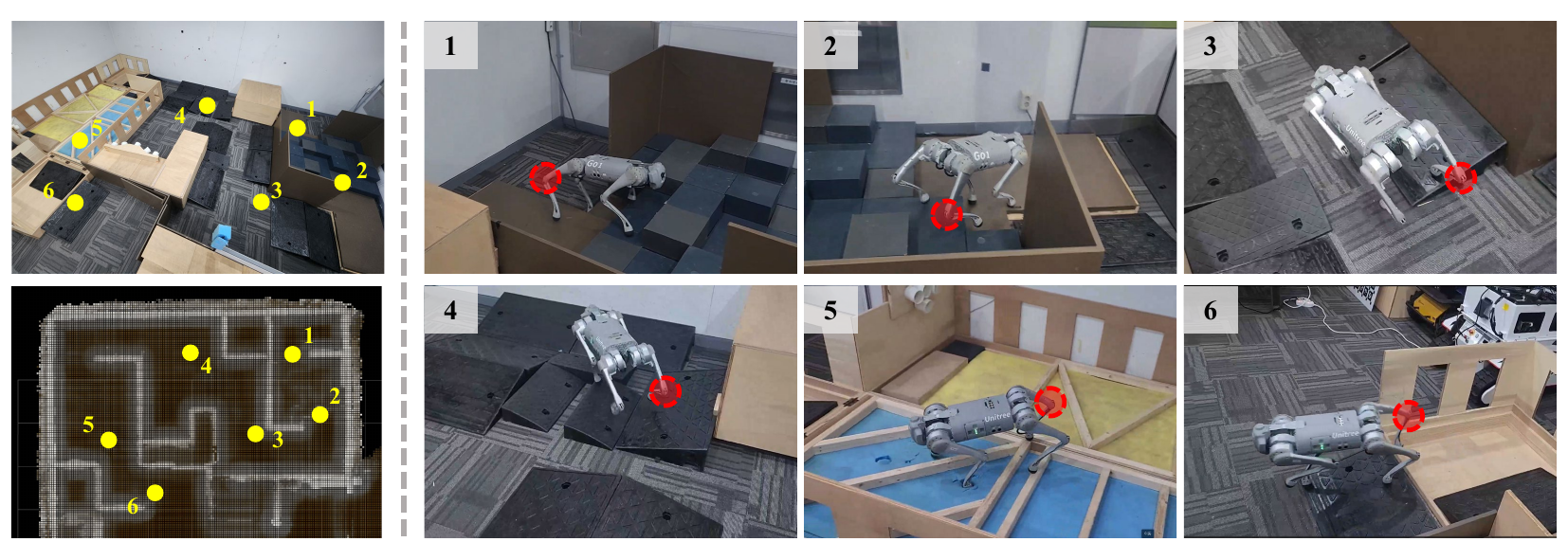}
	\caption{Our controller successfully overcame the indoor complex environment with the RR calf joint failure, i.e., weakened motor. (left column) 3D view and terrain map constructed using~\cite{b-tms} and (right column) the snapshots of the robot.}
	\label{fig:indoor_exp}
	\vsfig
\end{figure}

We also tested our proposed fault-tolerant locomotion controller in an indoor complex environment (Fig.~\ref{fig:indoor_exp}) to demonstrate its adaptability to various terrains. 
The indoor course consists of steps, bumps, and soft terrains, inspired by the QRC arena~\cite{qrc}.
Especially, the deformable soft terrain was not included in the training simulations.
Nevertheless, the robot successfully overcame the obstacles and maintained stable locomotion despite the faults in the rear leg joints.
This result demonstrates that our proposed method can adapt to out-of-distribution (OOD) states and maintain stable locomotion in these terrains. 

\section{Conclusion}
\label{sec:conclusion}

In this paper, we proposed a robust fault-tolerant locomotion controller, called \OurAlgorithm.
\OurAlgorithm\ is capable of handling various hardware failures that may occur in quadrupedal robots and enables the robot to walk through various unstructured environments.
Experimental results in a simulation environment show that \OurAlgorithm\ significantly outperforms the existing methods.
Furthermore, real-world experiments with actual robots in various unstructured environments
validated practical applicability and robustness of the proposed controller.
In future work, we plan to extend \OurAlgorithm\ to leverage an additional hardware such as a manipulator to enhance the robustness of the fault-tolerant controller when the robot is faced with various dynamic environments.
 
\bibliographystyle{URL-IEEEtrans}

\bibliography{URL-bib}

\end{document}